\documentclass[11pt,a4paper]{article}
\usepackage[hyperref]{acl2018}
\usepackage{times,latexsym,url,amsmath,graphicx,amsfonts,float,algorithm,algpseudocode,setspace}

\aclfinalcopy 

\title{Adversarial Learning of Task-Oriented Neural Dialog Models}

\author{Bing Liu \\
  Carnegie Mellon University \\
  Electrical and Computer Engineering \\
  {\tt liubing@cmu.edu} \\\And
  Ian Lane \\
  Carnegie Mellon University \\
  Electrical and Computer Engineering \\
  Language Technologies Institute \\
  {\tt lane@cmu.edu} \\}

\date{}
\begin{document}
\maketitle
\begin{abstract}
    In this work, we propose an adversarial learning method for reward estimation in reinforcement learning (RL) based task-oriented dialog models. Most of the current RL based task-oriented dialog systems require the access to a reward signal from either user feedback or user ratings. Such user ratings, however, may not always be consistent or available in practice. Furthermore, online dialog policy learning with RL typically requires a large number of queries to users, suffering from sample efficiency problem. To address these challenges, we propose an adversarial learning method to learn dialog rewards directly from dialog samples. Such rewards are further used to optimize the dialog policy with policy gradient based RL. In the evaluation in a restaurant search domain, we show that the proposed adversarial dialog learning method achieves advanced dialog success rate comparing to strong baseline methods. We further discuss the covariate shift problem in online adversarial dialog learning and show how we can address that with partial access to user feedback.
\end{abstract}

\section{Introduction}
    Task-oriented dialog systems are designed to assist user in completing daily tasks, such as making reservations and providing customer support. Comparing to chit-chat systems that are usually modeled with single-turn context-response pairs~\cite{li2016persona,serban2016building}, task-oriented dialog systems~\cite{young2013pomdp,williams2017hybrid} involve retrieving information from external resources and reasoning over multiple dialog turns. This makes it especially important for a system to be able to learn interactively from users.

    Recent efforts on task-oriented dialog systems focus on learning dialog models from a data-driven approach using human-human or human-machine conversations. Williams et al.~\shortcite{williams2017hybrid} designed a hybrid supervised and reinforcement learning end-to-end dialog agent. Dhingra et al.~\shortcite{dhingra2017towards} proposed an RL based model for information access that can learn online via user interactions. Such systems assume the model has access to a reward signal at the end of a dialog, either in the form of a binary user feedback or a continuous user score. A challenge with such learning systems is that user feedback may be inconsistent~\cite{su2016acl} and may not always be available in practice. Further more, online dialog policy learning with RL usually suffers from sample efficiency issue~\cite{suEtAl2017}, which requires an agent to make a large number of feedback queries to users.

    To reduce the high demand for user feedback in online policy learning, solutions have been proposed to design or to learn a reward function that can be used to generate a reward in approximation to a user feedback. Designing a good reward function is not easy~\cite{walker1997paradise} as it typically requires strong domain knowledge. El Asri et al.~\shortcite{el2014task} proposed a learning based reward function that is trained with task completion transfer learning. Su et al.~\shortcite{su2016acl} proposed an online active learning method for reward estimation using Gaussian process classification. These methods still require annotations of dialog ratings by users, and thus may also suffer from the rating consistency and learning efficiency issues.

    To address the above discussed challenges, we investigate the effectiveness of learning dialog rewards directly from dialog samples. Inspired by the success of adversarial training in computer vision~\cite{denton2015deep} and natural language generation~\cite{li2017adversarial}, we propose an adversarial learning method for task-oriented dialog systems. We jointly train two models, a generator that interacts with the environment to produce task-oriented dialogs, and a discriminator that marks a dialog sample as being successful or not. The generator is a neural network based task-oriented dialog agent. The environment that the dialog agent interacts with is the user. Quality of a dialog produced by the agent and the user is measured by the likelihood that it fools the discriminator to believe that the dialog is a successful one conducted by a human agent. We treat dialog agent optimization as a reinforcement learning problem. The output from the discriminator serves as a reward to the dialog agent, pushing it towards completing a task in a way that is indistinguishable from how a human agent completes it. 

    In this work, we discuss how the adversarial learning reward function compares to designed reward functions in learning a good dialog policy. Our experimental results in a restaurant search domain show that dialog agents that are optimized with the proposed adversarial learning method achieve advanced task success rate comparing to strong baseline methods. We discuss the impact of the size of annotated dialog samples to the effectiveness of dialog adversarial learning. We further discuss the covariate shift issue in interactive adversarial learning and show how we can address that with partial access to user feedback.

\section{Related Work}
    \textbf{Task-Oriented Dialog Learning} \hspace{3mm} Popular approaches in learning task-oriented dialog systems include modeling the task as a partially observable Markov Decision Process (POMDP)~\cite{young2013pomdp}. Reinforcement learning can be applied in the POMDP framework to learn dialog policy online by interacting with users~\cite{gavsic2013line}. 
    Recent efforts have been made in designing end-to-end solutions~\cite{williams2016end,Liu+2017,li2017end,liu2018dialogue} for task-oriented dialogs. Wen et al.~\shortcite{wenN2N16} designed a supervised training end-to-end neural dialog model with modularly connected components. Bordes and Weston~\shortcite{bordes2017} proposed a neural dialog model using end-to-end memory networks. These models are trained offline using fixed dialog corpora, and thus it is unknown how well the model performance generalizes to online user interactions. Williams et al.~\shortcite{williams2017hybrid} proposed a hybrid code network for task-oriented dialog that can be trained with supervised and reinforcement learning. Dhingra et al.~\shortcite{dhingra2017towards} proposed an RL dialog agent for information access. Such models are trained against rule-based user simulators. A dialog reward from the user simulator is expected at the end of each turn or each dialog. 

    \textbf{Dialog Reward Modeling} \hspace{3mm} Dialog reward estimation is an essential step for policy optimization in task-oriented dialogs. Walker et al.~\shortcite{walker1997paradise} proposed PARADISE framework in which user satisfaction is estimated using a number of dialog features such as number of turns and elapsed time. 
    Yang et al.~\shortcite{yang2012predicting} proposed a collaborative filtering based method in estimating user satisfaction in dialogs. Su et al.~\shortcite{su2015learning} studied using convolutional neural networks in rating dialog success. Su et al.~\shortcite{su2016acl} further proposed an online active learning method based on Gaussian process for dialog reward learning. These methods still require various levels of annotations of dialog ratings by users, either offline or online. On the other side of the spectrum, Paek and Pieraccini ~\shortcite{paek2008automating} proposed inferring a reward directly from dialog corpora with inverse reinforcement learning (IRL)~\cite{ng2000algorithms}. However, most of the IRL algorithms are very expensive to run~\cite{ho2016generative}, requiring reinforcement learning in an inner loop. This hinders IRL based dialog reward estimation methods to scale to complex dialog scenarios. 

    \textbf{Adversarial Networks} \hspace{3mm} Generative adversarial networks (GANs)~\cite{goodfellow2014generative} have recently been successfully applied in computer vision and natural language generation~\cite{li2017adversarial}. The network training process is framed as a game, in which people train a generator whose job is to generate samples to fool a discriminator. The job of a discriminator is to distinguish samples produced by the generator from the real ones. The generator and the discriminator are jointly trained until convergence. GANs were firstly applied in image generation and recently used in language tasks. Li et al.~\shortcite{li2017adversarial} proposed conducting adversarial learning for response generation in open-domain dialogs. 
    Yang et al.~\shortcite{yang2017improving} proposed using adversarial learning in neural machine translation. 
    The use of adversarial learning in task-oriented dialogs has not been well studied. Peng et al.~\shortcite{peng2017adversarial} recently explored using adversarial loss as an extra critic in addition to the main reward function based on task completion. This method still requires prior knowledge of a user's goal, which can be hard to collect in practice, in defining the completion of a task. Our proposed method uses adversarial reward as the only source of reward signal for policy optimization in addressing this challenge.

\section{Adversarial Learning for Task-Oriented Dialogs}
    In this section, we describe the proposed adversarial learning method for policy optimization in task-oriented neural dialog models. Our objective is to learn a dialog agent (i.e. the generator, $G$) that is able to effectively communicate with a user over a multi-turn conversation to complete a task. This can be framed as a sequential decision making problem, in which the agent generates a best action to take at every dialog turn given the dialog context. The action can be in the form of either a dialog act~\cite{henderson2013dialog} or a natural language utterance. We study on dialog act level in this work. Let $U_k$ and $A_k$ represent the user input and agent outputs (i.e. the agent act $a_k$ and the slot-value predictions) at turn $k$. Given the current user input $U_k$, the agent estimates the user's goal and select a best action $a_k$ to take conditioning on the dialog history. 

    In addition, we want to learn a reward function (i.e. the discriminator, $D$) that is able to provide guidance to the agent for learning a better policy. We expect the reward function to give a higher reward to the agent if the conversation it had with the user is closer to how a human agent completes the task. Output of the reward function is the probability of a given dialog being successfully completed. We train the reward function by forcing it to distinguish successful dialogs and dialogs conducted by the machine agent. At the same time, we also update the dialog agent parameters with policy gradient based reinforcement learning using the reward produced by the reward function. We keep updating the dialog agent and the reward function until the discriminator can no longer distinguish dialogs from a human agent and from a machine agent. In the subsequent sections, we describe in detail the design of our dialog agent and reward function, and the proposed adversarial dialog learning method.

\subsection{Neural Dialog Agent}
\label{fig:generator_model}
    \begin{figure}[t]
        \centering
        \includegraphics[width=\linewidth]{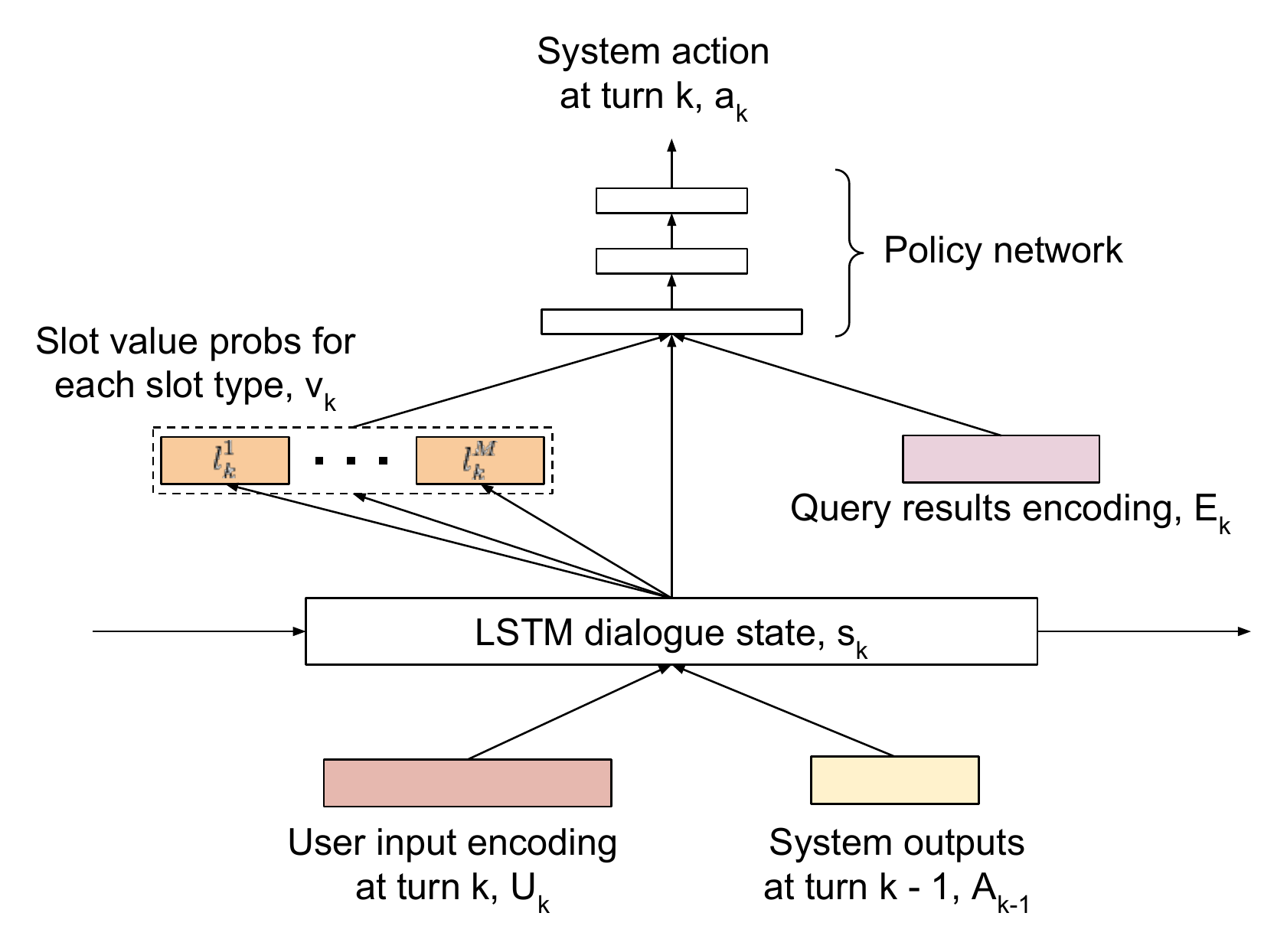}
        \vspace*{-2ex}
        \caption{{ Design of the task-oriented neural dialog agent. }}
        \label{fig:generator}
        \vspace*{-2ex}
    \end{figure}
    The generator is a neural network based task-oriented dialog agent. The model architecture is shown in Figure 1. The agent uses an LSTM recurrent neural network to model the sequence of turns in a dialog. At each turn, the agent takes a best system action conditioning on the current dialog state. A continuous form dialog state is maintained in the LSTM state $s_k$. At each dialog turn $k$, user input $U_k$ and previous system output $A_{k-1}$ are firstly encoded to continuous representations. The user input can either in the form of a dialog act or a natural language utterance. We use dialog act form user input in our experiment. The dialog act representation is obtained by concatenating the embeddings of the act and the slot-value pairs. If natural language form of input is used, we can encode the sequence of words using a bidirectional RNN and take the concatenation of the last forward and backward states as the utterance representation, similar to~\cite{yang2016hierarchical} and~\cite{Liu+2017}. With the user input $U_k$ and agent input $A_{k-1}$, the dialog state $s_k$ is updated from the previous state $s_{k-1}$ by:
        \begin{equation}
            s_k = \operatorname{LSTM_G}(s_{k-1}, \hspace{1mm} [U_k, \hspace{1mm} A_{k-1}])
        \end{equation}
    \textbf{Belief Tracking} \hspace{5mm}
    Belief tracking maintains the state of a conversation, such as a user's goals, by accumulating evidence along the sequence of dialog turns. A user's goal is represented by a list of slot-value pairs. The belief tracker updates its estimation of the user's goal by maintaining a probability distribution $P(l^{m}_k)$ over candidate values for each of the tracked goal slot type $m \in M$. With the current dialog state $s_k$, the probability over candidate values for each of the tracked goal slot is calculated by:
        \begin{equation}
            P(l^{m}_k \hspace{1mm} | \hspace{1mm} \mathbf{U}_{\le k}, \hspace{1mm} \mathbf{A}_{< k}) = \operatorname{SlotDist}_{m}(s_k)
        \end{equation}
    where $\operatorname{SlotDist}_{m}$ is a single hidden layer MLP with $\operatorname{softmax}$ activation over slot type $m \in M$.

    \textbf{Dialog Policy} \hspace{5mm}
    We model the agent's policy with a deep neural network. Following the policy, the agent selects the next action in response to the user's input based on the current dialog state. In addition, information retrieved from external resources may also affects the agent's next action. Therefore, inputs to our policy module are the current dialog state $s_k$, the probability distribution of estimated user goal slot values $v_{k}$, and the encoding of the information retrieved from external sources $E_{k}$. Here instead of encoding the actual query results, we encode a summary of the retrieved items (i.e. count and availability of the returned items). Based on these inputs, the policy network produces a probability distribution over the next system actions:
        \begin{equation}
            P(a_{k} \hspace{1mm} | \hspace{1mm} U_{\le k}, \hspace{1mm} A_{< k}, \hspace{1mm} E_{\le k}) = \operatorname{PolicyNet}(s_{k}, v_{k}, E_{k})
        \end{equation}
    where $\operatorname{PolicyNet}$ is a single hidden layer MLP with $\operatorname{softmax}$ activation over all system actions.    

\subsection{Dialog Reward Estimator}
\label{sec:discriminator_model}
    The discriminator model is a binary classifier that takes in a dialog with a sequence of turns and outputs a label indicating whether the dialog is a successful one or not. The logistic function returns a probability of the input dialog being successful. The discriminator model design is as shown in Figure 2. We use a bidirectional LSTM to encode the sequence of turns. At each dialog turn $k$, input to the discriminator model is the concatenation of (1) encoding of the user input $U_k$, (2) encoding of the query result summary $E_k$, and (3) encoding of agent output $A_k$. The discriminator LSTM output at each step $k$, $h_k$, is a concatenation of the forward LSTM output $\overrightarrow{h_k}$ and the backward LSTM output $\overleftarrow{h_k}$: $h_k = [\overrightarrow{h_k}, \overleftarrow{h_k}]$. 

    Once obtaining the discriminator LSTM state outputs $\{h_1, \dots, h_K\}$, we experiment with four different methods in combining these state outputs to generated the final dialog representation $d$ for the binary classifier:

    \textbf{BiLSTM-last} \hspace{3mm} Produce the final dialog representation $d$ by concatenating the last LSTM state outputs from the forward and backward directions: $d = [\overrightarrow{h_K}, \overleftarrow{h_1}]$

    \textbf{BiLSTM-max} \hspace{3mm} Max-pooling. Produce the final dialog representation $d$ by selecting the maximum value over each dimension of the LSTM state outputs. 

    \textbf{BiLSTM-avg} \hspace{3mm} Average-pooling. Produce the final dialog representation $d$ by taking the average value over each dimension of the LSTM state outputs. 

    \textbf{BiLSTM-attn} \hspace{3mm} Attention-pooling. Produce the final dialog representation $d$ by taking the weighted sum of the LSTM state outputs. The weights are calculated with attention mechanism:
        \begin{equation}
            d = \sum_{k=1}^{K}\alpha_{k}h_{k}
        \end{equation}        
        and
        \begin{equation}
        \begin{split}
            \alpha_{k} = \frac{\operatorname{exp}(e_{k})}{\sum_{t=1}^K\operatorname{exp}(e_{t})}, \hspace{3mm} e_{k} = g(h_{k})
        \end{split}
        \end{equation} 
        $g$ a feed-forward neural network with a single output node. 
    Finally, the discriminator produces a value indicating the likelihood the input dialog being a successful one:
        \begin{equation}
            D(d) = \sigma (W_{o}d + b_{o})
        \end{equation}
    where $W_o$ and $b_o$ are the weights and bias in the discriminator output layer. $\sigma$ is a logistic function. 

    \begin{figure}[t]
        \centering
        \includegraphics[width=\linewidth]{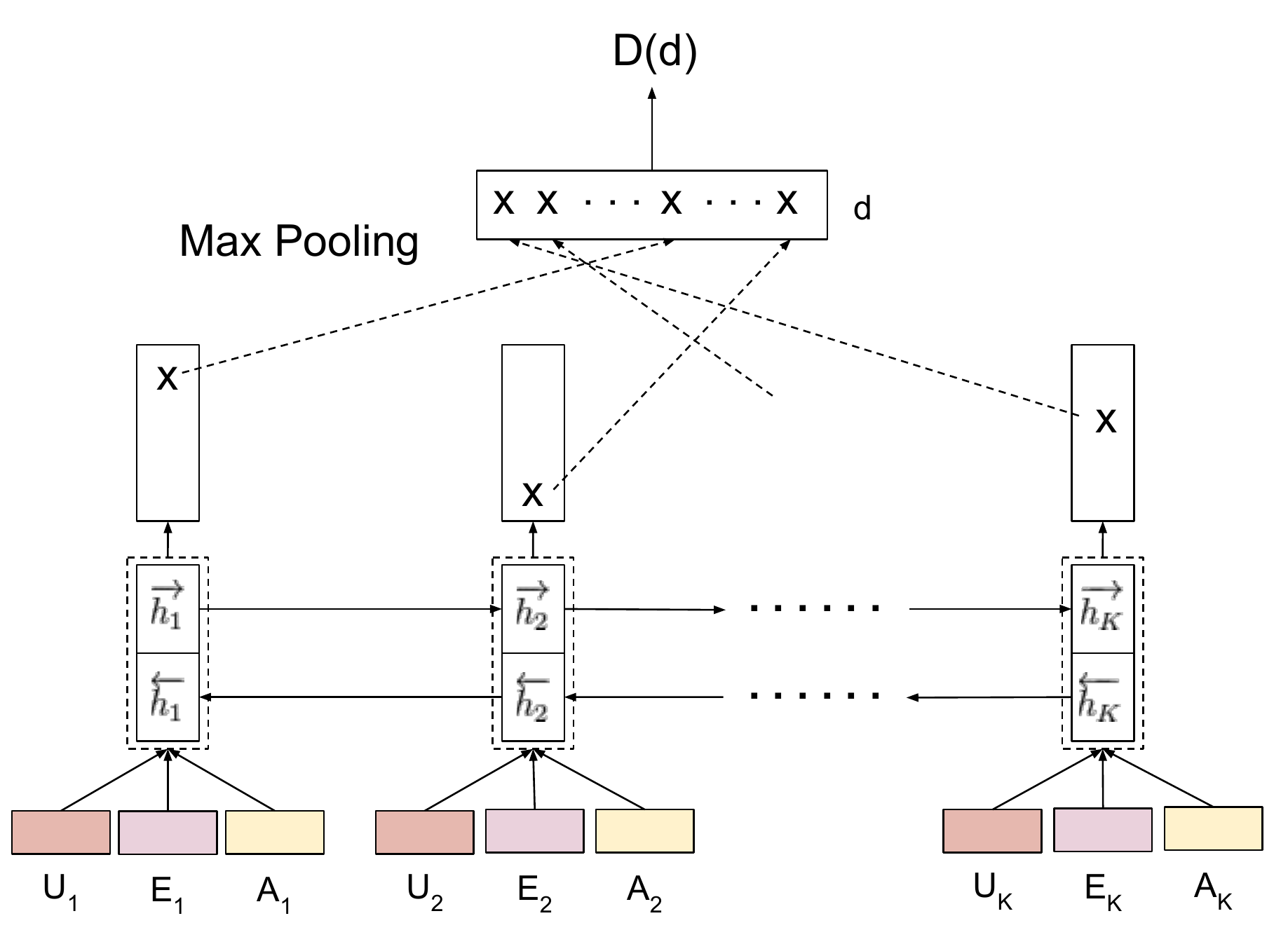}
        \vspace*{-2ex}
        \caption{{ Design of the dialog reward estimator: Bidirectional LSTM with max pooling. }}
        \label{fig:discriminator}
        \vspace*{-2ex}
    \end{figure}
    
\subsection{Adversarial Model Training}
    Once we obtain a dialog sample initiated by the agent and a dialog reward from the reward function, we optimize the dialog agent using REINFORCE~\cite{williams1992simple} with the given reward. The reward $D(d)$ is only received at the end of a dialog, i.e. $r_K = D(d)$. We discount this final reward with a discount factor $\gamma \in [0,1)$ to assign a reward $R_k$ to each dialog turn. The objective function can thus be written as $J_k(\theta_G) = \mathbb E_{\theta_G}\left[ R_k  \right] = \mathbb E_{\theta_G}\left[ \sum_{t=k}^{K} \gamma ^{t-k}r_{t} - V(s_k) \right]$, with $r_k = D(d)$ for $k=K$ and $r_k = 0$ for $k<K$. $V(s_k)$ is the state value function which serves as a baseline value. The state value function is a feed-forward neural network with a single-node value output. We optimize the generator parameter $\theta _G$ to maximize $J_k(\theta_G)$. With likelihood ratio gradient estimator, the gradient of $J_k(\theta_G)$ can be derived with:
        \begin{equation}
            \begin{split}
            \nabla  _{\theta _G} J_k(\theta _G) &= \nabla _{\theta _G} \mathbb E_{\theta _G}\left[ R_k \right] \\
            &= \sum_{a_{k} \in \mathcal{A}}  G(a_{k} | \cdot) \nabla _{\theta _G} \log G(a_{k} | \cdot) R_{k} \\
            &= \mathbb E_{\theta _G}\left[ \nabla _{\theta _G} \log G(a_{k} | \cdot) R_{k} \right]
            \end{split}
        \end{equation}
    where $G(a_{k} | \cdot) = G(a_{k} | s_{k}, v_{k}, E_{k}; \theta _G)$. The expression above gives us an unbiased gradient estimator. We sample agent action $a_k$ following a softmax policy at each dialog turn and compute the policy gradient. At the same time, we update the discriminator parameter $\theta_D$ to maximize the probability of assigning the correct labels to the successful dialog from human demonstration and the dialog conducted by the machine agent:
    \begin{equation}
    \begin{split}
        \nabla  _{\theta _D}\Big[\mathbb E_{d \sim \theta_{demo}} & \left[ log(D(d))  \right] + \\
        & \mathbb E_{d \sim \theta_G}\left[ log(1 - D(d))  \right]\Big]
    \end{split}
    \end{equation}
    We continue to update both the dialog agent and the reward function via dialog simulation or real user interaction until convergence.
    \begin{algorithm}[t]
    \caption{Adversarial Learning for Task-Oriented Dialog}
    \begin{algorithmic}[1]
    \State \textbf{Required:} dialog corpus $S_{demo}$, user simualtor $U$, generator $G$, discriminator $D$
    \State Pretrain a dialog agent (i.e. the generator) $G$ on dialog corpora $S_{demo}$ with MLE
    \State Simulate dialogs $S_{simu}$ between $U$ and $G$
    \State Sample successful dialogs $S_{(+)}$ and random dialogs $S_{(-)}$ from \{$S_{demo}$, $S_{simu}$\}
    \State Pretrain a reward function (i.e. the discriminator) $D$ with \mbox{$S_{(+)}$ and $S_{(-)}$ \hspace*{15mm} \Comment{eq 8}}
    \For{number of training iterations}
        \For{G-steps}
            \State Simulate dialogs $S_{b}$ between $U$ and $G$ 
            \State Compute reward $r$ for each dialog in \hspace*{11mm} $S_{b}$ with $D$ \Comment{eq 6}
            \State Update $G$ with reward $r$ \Comment{eq 7}
        \EndFor
        \For{D-steps}
        	\State Sample dialogs $S_{(b+)}$ from $S_{(+)}$
            \State Update $D$ with $S_{(b+)}$ and $S_{b}$ (with $S_{b}$ \hspace*{11mm} as negative examples) \Comment{eq 8}
        \EndFor
    \EndFor
    \end{algorithmic}
    \end{algorithm}
    
\section{Experiments}
\subsection{Dataset}
    We use data from the second Dialog State Tracking Challenge (DSTC2) \cite{henderson2014second} in the restaurant search domain for our model training and evaluation. We add entity information to each dialog sample in the original DSTC2 dataset. This makes entity information a part of the model training process, enabling the agent to handle entities during interactive evaluation with users. Different from the agent action definition used in DSTC2, actions in our system are produced by concatenating the act and slot types in the original dialog act output (e.g. ``$confirm(food=italian)$'' maps to ``$confirm\_food$''). The slot values (e.g. $italian$) are captured in the belief tracking outputs. Table 1 shows the statistics of the dataset used in our experiments.
        \begin{table}[ht]
          \label{tab:table_1}
          \centering
          \begin{tabular}{l r}
            \hline
            \# of train/dev/test dialogs           & 1612/506/ 1117     \\
            \# of dialog turns in average    & 7.88 \\ 
            \# of slot value options       &  \\
            \hspace{5mm} Area & 5 \\
            \hspace{5mm} Food & 91 \\
            \hspace{5mm} Price range & 3 \\
            \hline
          \end{tabular}     
        \caption{Statistics of DSTC2 dataset. }
        \end{table}

\subsection{Training Settings}
\label{sec:training_setting}
    We use a user simulator for our interactive training and evaluation with adversarial learning. Instead of using a rule-based user simulator as in many prior work~\cite{zhao2016towards,peng2017composite}, in our study we use a model-based simulator trained on DSTC2 dataset. We follow the design and training procedures of~\cite{liu2017iterative} in building the model-based simulator. The stochastic policy used in the simulator introduces additional diversity in user behavior during dialog simulation.

    Before performing interactive adversarial learning with RL, we pretrain the dialog agent and the discriminative reward function with offline supervised learning on DSTC2 dataset. We find this being helpful in enabling the adversarial policy learning to start with a good initialization. The dialog agent is pretrained to minimize the cross-entropy losses on agent action and slot value predictions. Once we obtain a supervised training dialog agent, we simulate dialogs between the agent and the user simulator. These simulated dialogs together with the dialogs in DSTC2 dataset are then used to pretrain the discriminative reward function. We sample 500 successful dialogs as positive examples, and 500 random dialogs as negative examples in pretraining the discriminator. During dialog simulation, a dialog is marked as successful if the agent's belief tracking outputs fully match the informable~\cite{henderson2013dialog} user goal slot values, and all user requested slots are fulfilled. This is the same evaluation criteria as used in \cite{wenN2N16} and \cite{liu2017iterative}. It is important to note that such dialog success signal is usually not available during real user interactions, unless we explicitly ask users to provide this feedback.

    During supervised pretraining, for the dialog agent we use LSTM with a state size of 150. Hidden layer size for the policy network MLP is set as 100. For the discriminator model, a state size of 200 is used for the bidirectional LSTM. We perform mini-batch training with batch size of 32 using Adam optimization method \cite{kingma2014adam} with initial learning rate of 1e-3. Dropout ($p=0.5$) is applied during model training to prevent the model from over-fitting. Gradient clipping threshold is set to 5. 

    During interactive learning with adversarial RL, we set the maximum allowed number of dialog turns as 20. A simulation is force to terminated after 20 dialog turns. We update the model with every mini-batch of 25 samples. Dialog rewards are calculated by the discriminative reward function. Reward discount factor  $\gamma$ is set as 0.95. These rewards are used to update the agent model via policy gradient. At the same time, this mini-batch of simulated dialogs are used as negative examples to update the discriminator.

\subsection{Results and Analysis}
    In this section, we show and discuss our empirical evaluation results. We first compare dialog agent trained using the proposed adversarial reward to those using human designed reward and using oracle reward. We then discuss the impact of discriminator model design and model pretraining on the adversarial learning performance. Last but not least, we discuss the potential issue of covariate shift during interactive adversarial learning and show how we address that with partial access to user feedback.

\subsubsection{Comparison to Other Reward Types}
    We first compare the performance of dialog agent using adversarial reward to those using designed reward and oracle reward on dialog success rate. Designed reward refers to reward function that is designed by humans with domain knowledge. In our experiment, based on the dialog success criteria defined in section \ref{sec:training_setting}, we \textit{design} the following reward function for RL policy learning:

    \begin{itemize}
      \item +1 for each informable slot that is correctly estimated by the agent at the end of a dialog.
      \item If ALL informable slots are tracked correctly, +1 for each requestable slot successfully handled by the agent.
    \end{itemize}
    
    In addition to the comparison to human designed reward, we further compare to the case of using oracle reward during agent policy optimization. Using oracle reward refers to having access to the final dialog success status. We apply a reward of +1 for a successful dialog, and a reward of 0 for a failed dialog. Performance of the agent using oracle reward serves as an upper-bound for those using other types of reward. For the learning with adversarial rewards, we use BiLSTM-max as the discriminator model. During RL training, we normalize the rewards produced by different reward functions.
        \begin{figure}[t]
          \centering
          \includegraphics[width=\linewidth]{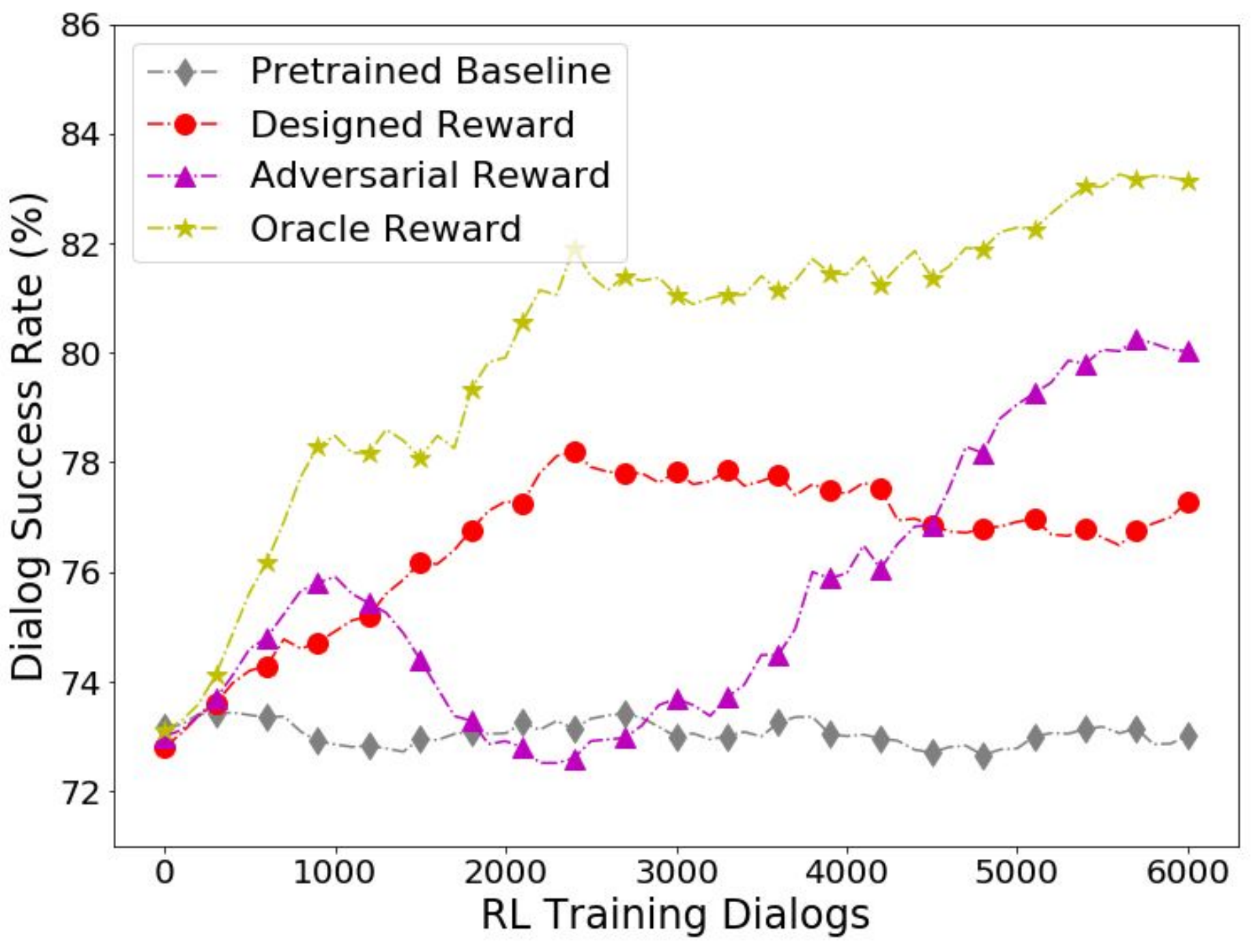}
          \vspace*{-5ex}
          \caption{ RL policy optimization performance comparing with adversarial reward, designed reward, and oracle reward. }
          \label{fig:diff_reward}
        \end{figure} 
    
    Figure \ref{fig:diff_reward} show the RL learning curves for models trained using different reward functions. The dialog success rate at each evaluation point is calculated by averaging over the success status of 1000 dialog simulations at that point. The pretrain baseline in the figure refers to the supervised pretraining model. This model does not get updated during interactive learning, and thus the curve stays flat during the RL training cycle. As shown in these curves, all the three types of reward functions lead to improved dialog success rate along the interactive learning process. The agent trained with designed reward falls behind the agent trained with oracle reward by a large margin. This shows that the reward designed with domain knowledge may not fully align with the final evaluation metric. Designing a reward function that can provide an agent enough supervision signal and also well aligns the final system objective is not a trivial task~\cite{popov2017data}. In practice, it is often difficult to exactly specify what we expect an agent to do, and we usually end up with simple and imperfect measures. In our experiment, agent using adversarial reward achieves a 7.4\% improvement on dialog success rate over the supervised pretraining baseline at the end of 6000 interactive dialog learning episodes, outperforming that using the designed reward (4.2\%). This shows the advantage of performing adversarial training in learning directly from expert demonstrations and in addressing the challenge of designing a proper reward function. Another important point we observe in our experiments is that RL agents trained with adversarial reward, although enjoy higher performance in the end, suffer from larger variance and instability on model performance during the RL training process, comparing to agents using human designed reward. This is because during RL training the agent interfaces with a moving target, rather than a fixed objective measure as in the case of using the designed reward or oracle reward. The model performance gradually becomes stabilized when both the dialog agent and the reward model are close to convergence.

\subsubsection{Impact of Discriminator Model Design}
    We study the impact of different discriminator model designs on the adversarial learning performance. We compare the four pooling methods described in section \ref{sec:discriminator_model} in producing the final dialog representation. Table 2 shows the offline evaluation results on 1000 simulated test dialog samples. Among the four pooling methods, max-pooling on bidirectional LSTM outputs achieves the best classification accuracy in our experiment. Max-pooling also assigns the highest probability to successful dialogs in the test set comparing to other pooling methods. Attention-pooling based LSTM model achieves the lowest performance across all the three offline evaluation metrics in our study. This is probably due to the limited number of training samples we used in pretraining the discriminator. Learning good attentions usually requires more data samples and the model may thus overfit the small training set. We observe similar trends during interactive learning evaluation that the attention-based discriminator leads to divergence of policy optimization more often than the other three pooling methods. Max-pooling discriminator gives the most stable performance during our interactive RL training.
        \begin{table}[ht]
          \label{tab:table_diff_discriminator}
          \centering
          \begin{tabular}{l c c c}
            \hline
            \textbf{}           & \textbf{Prediction}   & \textbf{Success}      & \textbf{Fail}     \\
            \textbf{Model}      & \textbf{Accuracy}     & \textbf{Prob.}        & \textbf{Prob.}     \\
            \hline
            BiLSTM-last         & 0.674                 & 0.580                             & 0.275 \\
            BiLSTM-max          & \textbf{0.706}        & \textbf{0.588}                    & 0.272 \\
            BiLSTM-avg          & 0.688                 & 0.561                             & \textbf{0.268} \\
            BiLSTM-attn         & 0.652                 & 0.541                             & 0.285 \\
            \hline
          \end{tabular}     
        \caption{Performance of different discriminator model design, on prediction accuracy and probabilities assigned to successful and failed dialogs.}
        \end{table}

\subsubsection{Impact of Annotated Dialogs for Discriminator Training}
        \begin{figure}[t]
          \centering
          \includegraphics[width=\linewidth]{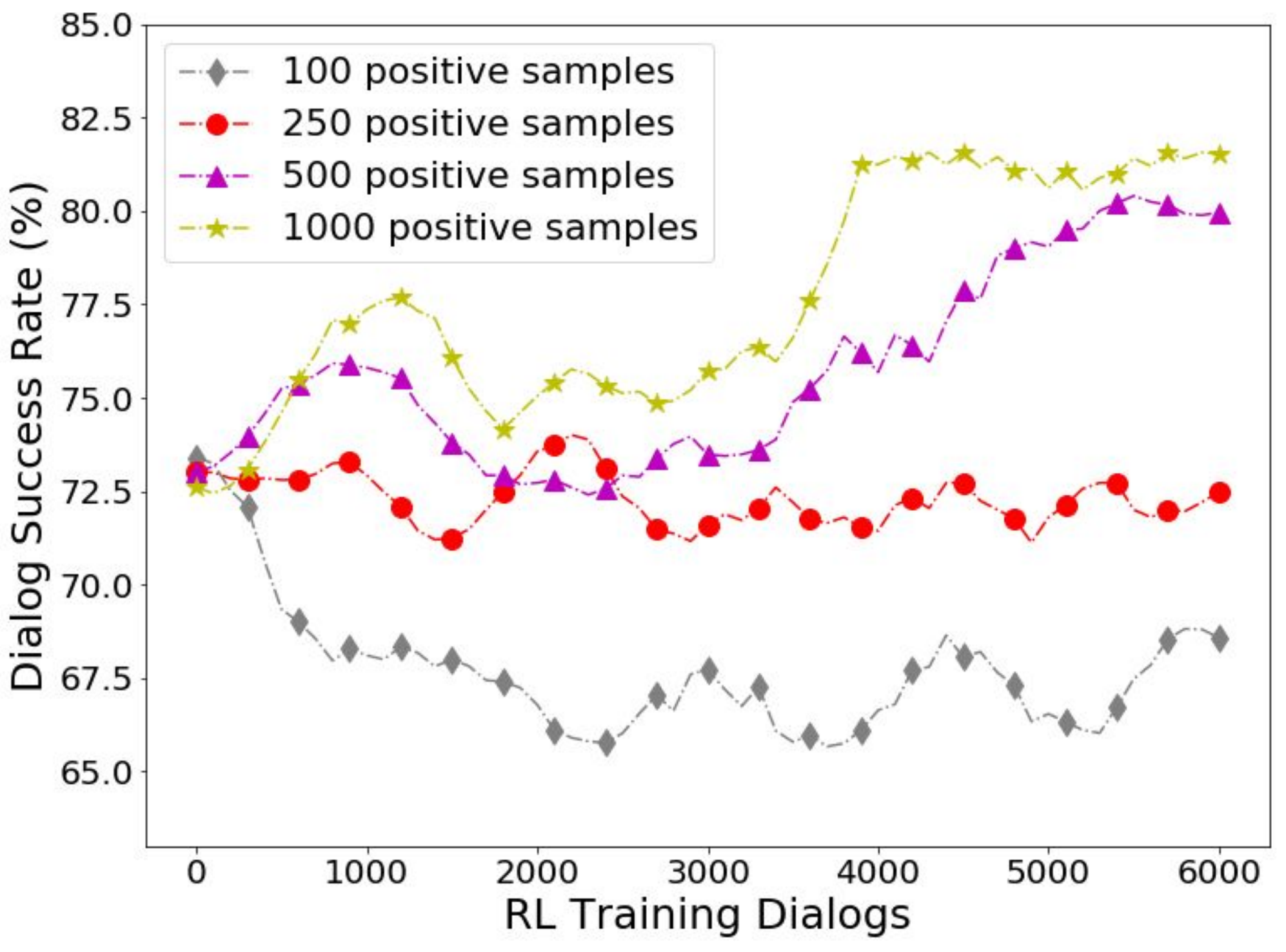}
          \vspace*{-4ex}
          \caption{Impact of discriminator training sample size on RL dialog learning performance. }
          \label{fig:diff_pretrain}
        \end{figure}
    Annotating dialogs for model training requires additional human efforts. We investigate the impact of the size of the annotated dialog samples on discriminator model training. The amount of annotated dialogs required for learning a good discriminator depends mainly on the complexity of a task. Given the rather simple nature of the slot filling based DSTC2 restaurant search task, we experiment with annotating 100 to 1000 discriminator training samples. We use BiLSTM-max discriminator model in these experiments. The adversarial RL training curves with different levels of discriminator training samples are shown in Figure \ref{fig:diff_pretrain}. As these results illustrate, with 100 annotated dialogs as positive samples for discriminator training, the discriminator is not able to produce dialog rewards that are useful in learning a good policy. Learning with 250 positive samples does not lead to concrete improvement on dialog success rate neither. With the growing number of annotated samples, the dialog agent becomes more likely to learn a better policy, resulting in higher dialog success rate at the end of the interactive learning sessions.

\subsubsection{Partial Access to User Feedback}
\begin{figure}[t]
          \centering
          \includegraphics[width=\linewidth]{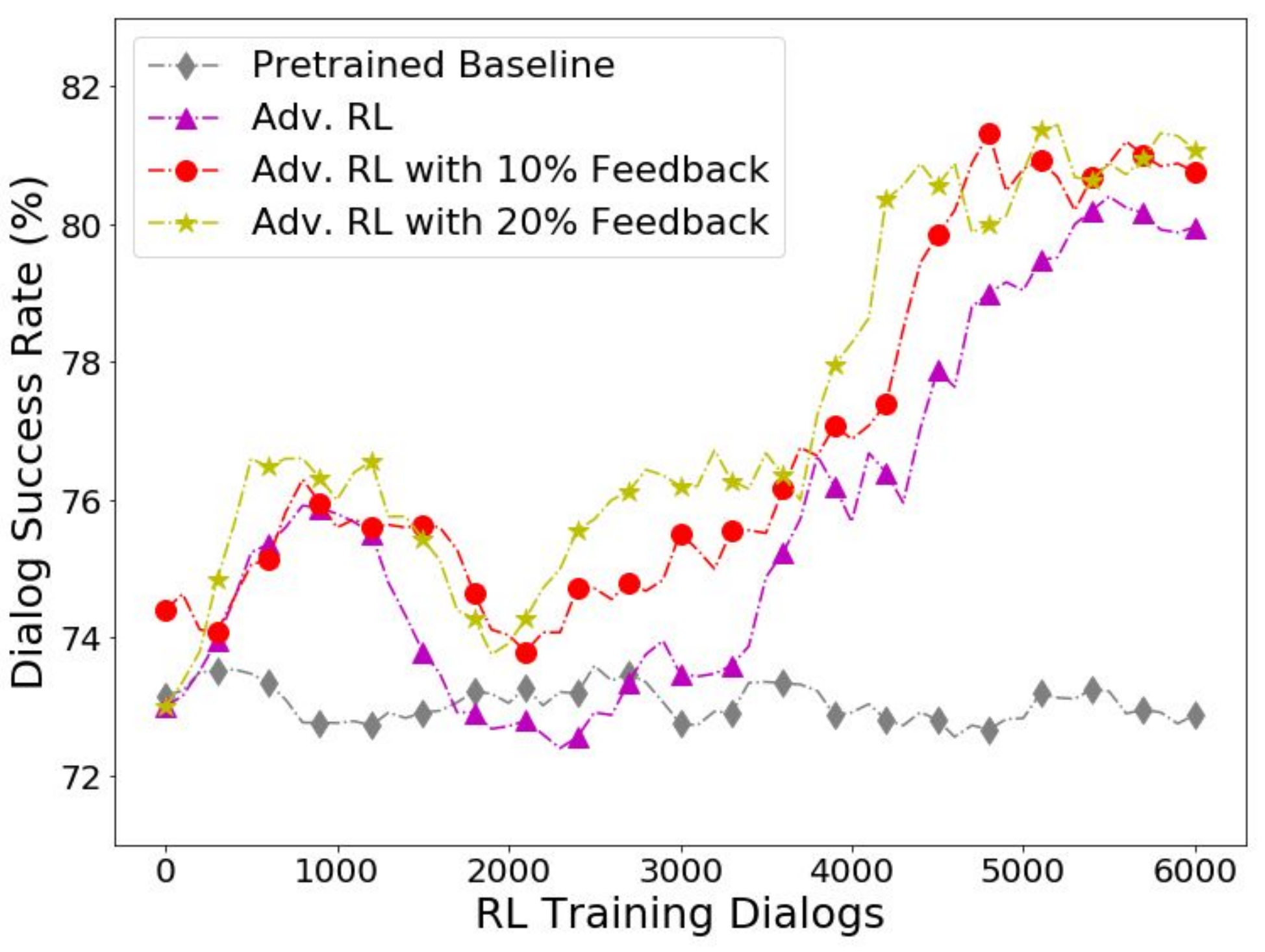}
          \vspace*{-4ex}
          \caption{Addressing covariate shift in online adversarial dialog learning with partial access to user feedback.}
          \label{fig:diff_feedback}
        \end{figure}

    A potential issue with RL based interactive adversarial learning is the covariate shift~\cite{ross2010efficient,ho2016generative} problem. Part of the positive examples for discriminator training are generated based on the supervised pretraining dialog policy before the interactive learning stage. During interactive RL training, the agent's policy gets updated. The newly generated dialog samples based on the updated policy may be equally good comparing to the initial set of positive dialogs, but they may look very different. In this case, the discriminator is likely to give these dialogs low rewards as the pattern presented in these dialogs is different to what the discriminator is initially trained on. The agent will thus be discouraged to produce such type of successful dialogs in the future with these negative rewards. To address such covariate shift issue, we design a DAgger~\cite{ross2011reduction} style imitation learning method to the dialog adversarial learning. We assume that during interactive learning with users, occasionally we can receive feedback from users indicating the quality of the conversation they had with the agent. We then add those dialogs with good feedback as additional training samples to the pool of positive dialogs used in discriminator model training. With this, the discriminator can learn to assign high rewards to such good dialogs in the future. In our empirical evaluation, we experiment with the agent receiving positive feedback 10\% and 20\% of the time during its interaction with users. The experimental results are shown in Figure \ref{fig:diff_feedback}. As illustrated in these curves, the proposed DAgger style learning method can effectively improve the dialog adversarial learning with RL, leading to higher dialog success rate.
\section{Conclusions}
    In this work, we investigate the effectiveness of applying adversarial training in learning task-oriented dialog models. The proposed method is an attempt towards addressing the rating consistency and learning efficiency issues in online dialog policy learning with user feedback. We show that with limited number of annotated dialogs, the proposed adversarial learning method can effectively learn a reward function and use that to guide policy optimization with policy gradient based reinforcement learning. In the experiment in a restaurant search domain, we show that the proposed adversarial learning method achieves advanced dialog success rate comparing to baseline methods using other forms of reward. We further discuss the covariate shift issue during interactive adversarial learning and show how we can address it with partial access to user feedback.

\bibliography{acl2018}
\bibliographystyle{acl_natbib}

\end{document}